\documentclass[conference]{IEEEtran}
\IEEEoverridecommandlockouts

\usepackage{amsmath,amsfonts,amssymb,amsthm,psfrag,enumerate}
\usepackage{graphicx}

\usepackage{algorithm}
\usepackage{framed}
\usepackage{algpseudocode}
\usepackage{mathrsfs}

\algdef{SE}[SUBALG]{Indent}{EndIndent}{}{\algorithmicend\ }%
\algtext*{Indent}
\algtext*{EndIndent}

\newcommand{\R}{\mathbb{R}}

\usepackage[framemethod=tikz]{mdframed}

\definecolor{mygray}{RGB}{248,248,250}
\newtheoremstyle{mystyle}
  {\topsep}
  {\topsep}
  {}
  {}
  {\bfseries}
  {.}
  {5pt plus 1pt minus 1pt}
  {{\color{black}\thmname{#1}~\thmnumber{#2}}\thmnote{\,--\,#3}}%
\theoremstyle{mystyle}
\newmdtheoremenv[%
  backgroundcolor=mygray,%
  linecolor=black,%
  leftmargin=0pt,%
  innerleftmargin=5pt,%
  innerrightmargin=5pt,%
  ]{probx}{Problem}

\title{\LARGE \bf
Fed-BEV: A Federated Learning Framework for Modelling Energy Consumption of Battery Electric Vehicles}

\author{Mingming Liu
\thanks{M. Liu is with the School of Electronic Engineering, Dublin City University, Dublin, Ireland. Email: {\tt mingming.liu@dcu.ie}. The author acknowledges the support from the school at DCU to carry out this project.}}

\begin{document}

\maketitle
\thispagestyle{empty}
\pagestyle{empty}

\begin{abstract}
Recently, there has been an increasing interest in the roll-out of electric vehicles (EVs) in the global automotive market. Compared to conventional internal combustion engine vehicles (ICEVs), EVs can not only help users reduce monetary costs in their daily commuting, but also can effectively help mitigate the increasing level of traffic emissions produced in cities. Among many others, battery electric vehicles (BEVs) exclusively use chemical energy stored in their battery packs for propulsion. Hence, it becomes important to understand how much energy can be consumed by such vehicles in various traffic scenarios towards effective energy management. To address this challenge, we propose a novel framework in this paper by leveraging the federated learning approaches for modelling energy consumption for BEVs (Fed-BEV). More specifically, a group of BEVs involved in the Fed-BEV framework can learn from each other to jointly enhance their energy consumption model. We present the design of the proposed system architecture and implementation details in a co-simulation environment. Finally, comparative studies and simulation results are discussed to illustrate the efficacy of our proposed framework for accurate energy modelling of BEVs. 
\end{abstract}


\section{Introduction}

Most recently, the Irish government has published new climate law which commits Ireland to net-zero carbon emissions by 2050 \cite{Ireland2050}. To achieve this target, electric vehicles, hybrid electric vehicles as well as many other green transportation tools will certainly play an important role to actively reduce the amount of pollutants produced in the traffic sector \cite{gu2021fair}. Among others, battery electric vehicles (BEVs) purely use chemical energy stored in their battery packs for propulsion, and thus it is important to quantitatively evaluate how much energy can be consumed by BEVs in different real-world traffic scenarios to avoid unsafe battery operations and to develop proper controlling algorithms for proactive maintenance strategies \cite{zhang2014battery}.

In order to build an accurate energy consumption model for BEVs, broadly speaking, two different design strategies can be deployed, namely analytical methods and data-driven methods. Typically, a classic analytical method works in the following three steps: (1) modelling and analysing the behaviour of each electrical and mechanical component of the vehicle; (2) linking and assembling various mechatronic components as subsystems/functional blocks as part of a whole vehicular system; and (3) simulating and testing the dynamics of a vehicle under certain circumstances to evaluate the performance metrics of interest, such as battery power losses, mechanical losses of motors, among others. Along this line, a large body of work has been found in the literature, see papers \cite{zhang2014battery, wang2015electric, miri2021electric, kaloko2011design, yang2014electric} for more details.

Compared to an analytical method, a data-driven method does not intend to capture the device-level details of each component in a vehicle, instead it aims to collect input-output data flows from sensors/components of a vehicle in different working conditions for further analysis. Through data mining and machine learning techniques, the collected data can be used to better predict the performance of a vehicle in both known and unknown working conditions. For instance, authors in \cite{qi2018data} proposed an energy consumption estimation model, using decision-tree models,  principle component analysis as well as neural networks, to fit real data collected from a 2013 Nissan Leaf. The trained model has shown an outstanding performance compared to the baseline models adapted from existing analytical models. Compared to the traditional analytical based methods for energy consumption modelling,  data-driven based design methods are emerging techniques, selected examples of work in the area can be found in \cite{yao2019vehicle, zhang2020energy}.

Clearly, one advantage of the analytical-based design methods is that it models all details for each electrical and mechanical component of the vehicle as well as their operational logic as part the whole system, and thus it allows a researcher to deeply explore the fundamental mechanism of the vehicle. However, since the modelling process is mainly based on the physical and mathematical properties of the specific vehicular model of interest, inevitably it has to involve lots of domain knowledge of the vehicle before further analysis and simulation works can be carried out. From this perspective, such a design method is not time-efficient to analyse and understand a new vehicular model. In contrast, a data-driven based method is useful as the analysis can be implemented based on the observable data from the vehicle, which in principle does not require a deep understanding of the vehicular system itself, and thus it is time-efficient to explore a novel model. However, this design method is largely based on the effectiveness of parameter estimation methods and the machine learning models, and thus it is mostly appealing for system-level analysis.

Note that for data-driven approaches the observable dataset is usually collected through a fixed set of vehicular configurations with a limited number of working conditions/driving cycles from specific real world scenarios. As a result, a trained learning model may not fully capture all operational characteristics of the vehicle in many other working conditions. For instance, a trained model based on the dataset collected from a vehicle which mostly travels out during warm weather with low loading ratio in a mountainous area will be most likely not applicable to a same type of vehicle which often travels out during cold weather with high loading ratio in a flat area. Thus, it becomes important to build a machine learning model which can not only appreciate this diversity but also can mitigate bias for accurately predicting energy consumption of the vehicle in different context. Clearly, the first step to deal with this challenge is to collect a large dataset from a group of vehicles sharing the same production model under various working conditions. Nevertheless, collecting and transmitting heterogeneous local data for model training and updating in an online centralized manner is not preferable as users' privacy can be easily compromised. In fact, a recent paper \cite{liu2020privacy} explicitly points out that acquisition of massive user data is not possible in real world applications. For instance, Tesla motors leaked the vehicle's location information when the vehicle's GPS data was used for model training purposes, which would inevitably lead to many privacy concerns to the owner of the vehicle. Such a security risk can be further escalated for users in EU when strict GDPR regulations need to be complied. Furthermore, although anonymization can be applied to the dataset in an offline fashion, this process can be very time-consuming and subject to processing errors. Thus, an online privacy-aware update of model is desirable, and such a design consideration has been ignored in most prior works.

Given this context, our objective in this paper is to leverage the benefits of both design methods aforementioned and integrate them into a privacy-aware model training framework. To this end, we borrow some key ideas from federated learning, a recently well-known machine learning paradigm that is able to learn a shared model from decentralized training data holding by each individual agent without revealing each local data to a central computing node. Our contributions in this work can be summarized as follows: 

\begin{itemize}
	\item[A.] We propose an end-to-end model learning framework for BEVs involved with federated learning, namely Fed-BEV. The framework consists of data generation, processing, model learning and deployment in a closed-loop setup. 
	\item[B.] We present a comprehensive comparative study using the proposed framework for different settings of interest, and we demonstrate the merits of our proposed framework through results from a co-simulation environment.
	\item[C.] To the best of the author's knowledge, this is the \textit{first time} that the federated learning based approach is applied to address the energy modelling issue for BEVs.
\end{itemize}

The remainder of the paper is organized as follows. Section \ref{statement} introduces the preliminaries and presents the problem statement of our work. Section \ref{architecture} details the system model and architecture for the energy consumption modelling problem, and discusses how federated learning techniques can be applied to solve the problem of our interest. Section \ref{Simulation} elaborates the simulation setup and illustrates our simulation results. Finally, Section \ref{Conclusion} highlights the key findings in the paper and outlines some directions for future research.

\section{Problem Statement} \label{statement}

We consider a scenario where a number of BEVs having the same production model are travelling in a city. We assume that these vehicles are willing to participate into the Fed-BEV programme with a common goal to collaboratively enhance capabilities of the model for energy consumption prediction. To do this, it is assumed that each vehicle in the programme is equipped with an onboard  communication/computing unit, e.g. an onboard vehicular computer system, such that the data flows collected from the local sensors of the vehicle can be stored, processed and trained using the onboard unit. Most importantly, it is required that the locally trained model can communicate with a nearby infrastructure, e.g. a road side unit (RSU) with a mobile edge computing (MEC) server, and such an infrastructure can occasionally send some updated information back to the vehicle for model updates. We note that this process can be easily established through current available communication infrastructures, and this process may be asynchronized, i.e. no real-time feedback is required from the infrastructure. The key fact is that although a synchronized update process is fast and preferable for all vehicles, there is no guarantee to the availability of all vehicles for such an update in a practical scenario. We will show later this situation can still be easily managed thanks to federated learning.

Given this context, we now formulate the energy modelling problem for BEVs as follows. Let $N$ be the total number of BEVs involved in the programme. Let $\underline{\textrm{N}}: = \left\lbrace 1,2,3, \dots, N \right\rbrace$ be the set for indexing these BEVs. For a given day, we assume that a vehicle $i \in \underline{\textrm{N}}$ travels a certain number of trips denoted by $n_i$. Each trip $T_i^{j}, ~\forall j \in \left\lbrace 1, 2, ..., n_i \right\rbrace$ is essentially a data matrix consisting of several time-series sequences for features of interest. Mathematically, we define:

\begin{equation*}
T_i^{j}: =  \left[ f_{i_0}^{j},  f_{i_1}^{j},  f_{i_2}^{j}, \ldots, f_{i_p}^{j}\right]
\end{equation*}

\noindent where 
\begin{equation*}
f_{i_h}^{j}: = \left[ f_{i_h}^{j}(1), f_{i_h}^{j}(2), \ldots, f_{i_h}^{j}(l_j) \right]^{\textrm{T}}, \forall h = \left\lbrace 0, 1, \ldots, p \right\rbrace.
\end{equation*}

Each $f_{i_h}^{j}(k)$ presents the time series data capturing the $h$'th feature of vehicle $i$ during the trip $j$ at time slot $k$ where the duration of the trip is set by $l_j$ in standard time units. For instance, the first feature $h=0$ can represent the time stamp information of the vehicle during the trip, the second feature $h=1$ can represent the speed profile of the vehicle during the trip, the third feature $h=2$ can contain the road slope information along the trip, and the last feature $h=p$ can represent the energy consumption data of the vehicle during the trip. Thus, there are a maximum number of $p$ input features can be used for the model training purpose, and the last feature $h=p$ for vehicle's energy consumption can be used as label. Given this, we define the daily historical trip records of the $i$'th vehicle as a collection of all trips during the day as follows:
\begin{equation*}
D^i: = \left[ T_i^{1}, T_i^{2}, \ldots, T_i^{n_i} \right]^{\textrm{T}} : = \left[D_0^i, D_1^i, \ldots, D_h^i, \ldots, D_p^i \right] 
\end{equation*}

\noindent where $D_h^i \in \R^{\sum_{j=1}^{n_i}l_j}$ is a column vector of $D^i$ that only the $h$'th feature of the trip data $T_i^j, \forall j $ is included. Thus, the input feature set for $i$'th vehicle's model training can be defined as: 

\begin{equation*}
F^i : = \left[D_{0}^i, D_{1}^i, \ldots, D_{p-1}^i \right]
\end{equation*}

\noindent and the output label data, i.e. the energy consumption data of the vehicle $i$, can be defined as:

\begin{equation*}
E^i: =  D_{p}^i 
\end{equation*}

Finally, for a given time $k$, we define the past $m$ observations from the feature set and the label set as $F_{k-m+1:k}^i$ and $E_{k-m+1:k}^i$ respectively. Note that $E_{k-m+1:k}^i$ is a vector with dimension $m$ which represents the energy consumption at every given time point within the past $m$ steps in standard time units. Here, instead of looking into the point-wise data of the vector, we consider the overall energy consumption data during the past $m$ steps as:

\begin{equation*}
\hat{E}_{k-m+1:k}^i := \textbf{1}^{\textrm{T}} E_{k-m+1:k}^i ~ \forall k \in \mathcal{K}, 
\end{equation*}

\noindent where $\textbf{1} \in \R^{m}$ is the column vector with all entries equal to 1, and $\mathcal{K}: =\left\lbrace m , m+1, \dots, \sum_{j=1}^{n_i}l_j \right\rbrace$ is a feasible indexing set. \\

Given the notation above, a local model training process for vehicle $i$ can be defined to find a local hypothesis function $H_i(.)$ which is able to address the following problem:
\begin{equation}\label{localopt}
\begin{aligned}
\min_{H_i} \quad & \sum_{k \in \mathcal{K}} | \hat{E}_{k-m+1:k}^i - \widetilde{E}_{k-m+1:k}^i | \\
\textrm{s.t.} \quad & \widetilde{E}_{k-m+1: k}^i = H_i(F_{k-m+1:k}^i) \\
\end{aligned}
\end{equation}

\textbf{Comment:} The problem \eqref{localopt} is formulated so that the overall energy consumption in a time window with fixed size $m$ can be predicted and the bias between the ground truth $\hat{E}_{k-m+1:k}^i$ and the \textit{predicted scalar} $\widetilde{E}_{k-m+1:k}^i$ can be minimized when $H_i$ is fine tuned. It is important to note that each local model training process is an essential step to obtain an energy model whilst preserving training data locally. The key difference between our work and the previous works is that we not only train local model $H_i$ but also we investigate how federated learning techniques and architecture can be applied to further enhance the capabilities of $H_i$ in this specific use case. For this purpose, we now present the system architecture and its implementation for Fed-BEV in the following section.

\section{System Model and Architecture} \label{architecture}

\begin{figure}[htbp]
	\begin{center}
		{\includegraphics[width=0.47\textwidth]{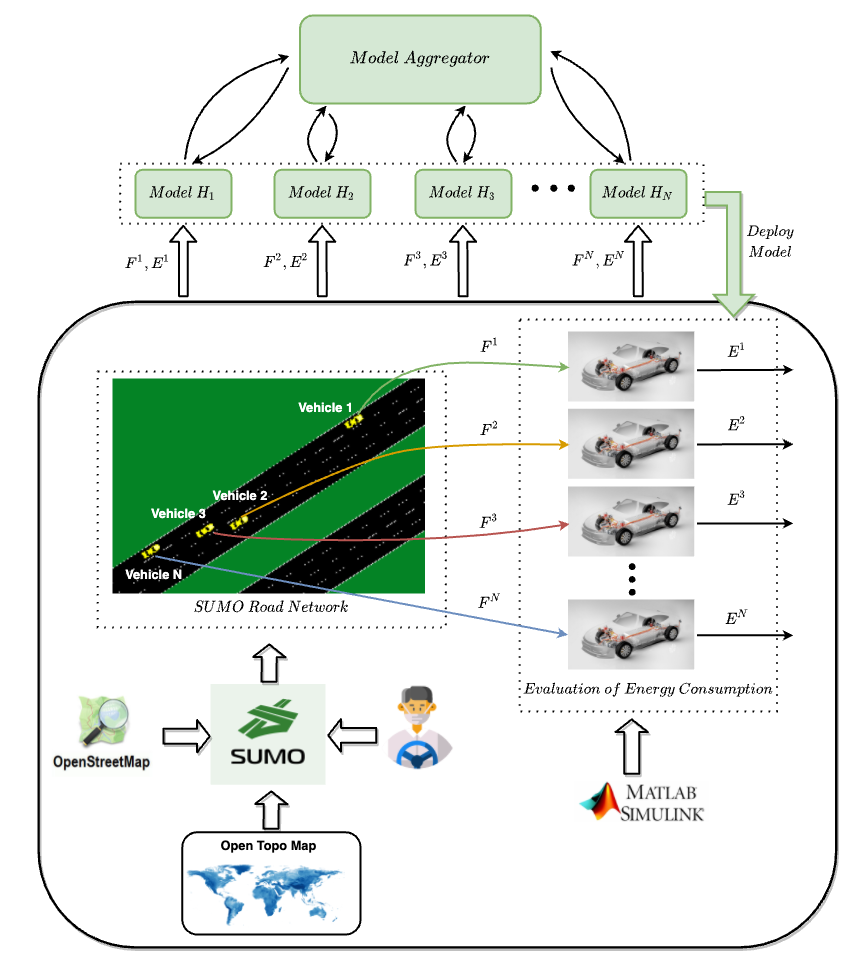}}
		\caption{The proposed system architecture for Fed-BEV.}
		\label{sys_architecture}
	\end{center}
\end{figure}

The proposed system architecture for Fed-BEV is shown in Fig. \ref{sys_architecture}. More specifically, the proposed system architecture consists of three key functional blocks jointly working together, namely 1. a traffic simulator module; 2. an energy evaluation module; and 3. a federated learning module including both local model generators and a model aggregator. Given this proposed architecture, our system operates as follows. 

\begin{itemize}
	\item[1.] The mobility data of vehicles is produced from a well-known traffic simulator, i.e. SUMO \cite{behrisch2011sumo} for any scenario of interest. In particular, SUMO takes account of three inputs, including a. OpenStreetMap \cite{haklay2008openstreetmap}, which can provide the information of road networks for vehicles travelling in the city; b. Drivers' travelling behaviours, which contain the origin-destination pairs of drivers in the road network, the specific trajectory information of trips undertaken, and how aggressive a driver steers his/her vehicle; and finally c. OpenTopoMap \footnote{https://www.opentopodata.org/}, which can be used to provide the topography information of the road networks including the elevation and potentially the slope information of a road segment that the driver is currently on. The output of SUMO can provide motional and contextual feature data of all vehicles in the road networks under the test, this information is then passed through an energy evaluation module in the following step. 
		
	\item[2.] The energy evaluation module is deployed here which ingests the motional data of vehicles generated from the first step to produce energy consumption data of vehicles for further model training process. The specific vehicular model for BEV is integrated through Matlab/Simulink \footnote{https://www.mathworks.com/products/simulink.html} to effectively evaluate the energy consumption for a BEV involved in the Fed-BEV. 
	
	\item[3.] A global model based on a centralized federated learning method is trained in an iterative manner by leveraging each local model trained through local vehicular data collected from the first two steps. 
 
	\item[4.] After the convergence of the federated learning algorithm in Step 3, the trained global model is deployed to each local vehicle to enhance accuracy of the energy consumption model.
	
\end{itemize}

	In the following we present details of each functional block in the proposed system architecture. 
	
\subsection{The Vehicular Model}

The objective of a vehicular model is to evaluate the energy consumption of a vehicle in the network as per the mobility pattern of the vehicle. Instead of using real-world dataset from production vehicles, which is usually commercial and very hard to obtain from open access data sources, we consider a classic BEV model constructed through Matlab/Simulink inspired by the works in \cite{mohan2014advanced}. We note that the purpose of this section is to give a quick review on the model used in the work, and demonstrate that how easy such a model can be integrated into our architecture to provide the required information for our model training. In other words, our intention is to well capture the input-output data flow for \textit{any given vehicular model as a black box}, where the input-output relation can be highly nonlinear in nature. We note that this simulation based module can be easily replaced when realistic data sources are provided through proper interfaces. 

\begin{figure}[htbp]
	\begin{center}
		\hspace{-0.5cm}
		{\includegraphics[width=0.5\textwidth, height=1.8in]{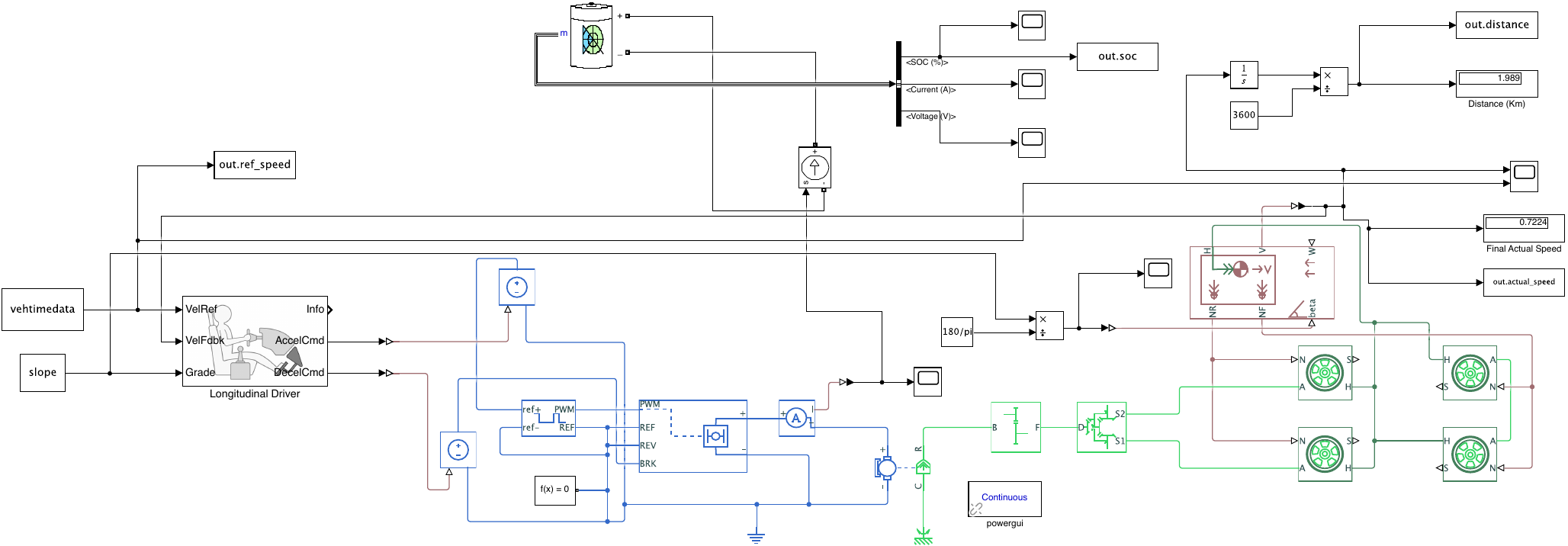}}
		\caption{The structure of our used vehicular model for BEVs.}
		\label{BEV_Model}
	\end{center}
\end{figure}

The structure of the vehicular model is presented in Fig. \ref{BEV_Model}. Broadly speaking, the model consists of four key functional blocks, i.e. mobility data of the vehicle collected from SUMO, speed controller and electric motor, battery supply and power converter, as well as the vehicular body, connected in a closed-loop setup. More specifically, the vehicle body which locates at the rightmost of the Fig. \ref{BEV_Model} attaches four wheels which are then connected to an electric motor through a transmission system involved with a simple gear train and a differential block. The vehicle body takes inputs from the reference speed and the road slope profiles from SUMO to generate required load and torque demand which need to be fulfilled by the battery supply through the power converter. On the one side, the power converter regulates the required power from the battery packs through PWM signals (the H-bridge block in blue) which are generated based on the required acceleration/de-acceleration values calculated using the difference of the reference speed profile and the resultant actual/feedback speed profile of the vehicle. On the other side, during braking, the power converter is also able to convert the excess kinetic energy of the vehicle to electric power through PWM signals and then revert such electrical energy to the battery packs. Finally, the leftmost part of the Fig. \ref{BEV_Model} is a parametric longitudinal speed tracking controller which is used for generating normalized acceleration and braking commands based on reference and feedback velocities of the vehicle body. The controller subsystem contains a PI controller with a nominal speed set to 100km/h. Finally, we note that the structure of vehicle model remains the same for each BEV, but some parameter configurations may differ from each other which will be discussed in Section \ref{Simulation}.

\subsection{Stacked-LSTM Architecture}

\begin{figure}[htbp]
	\begin{center}
		\hspace{-0.5cm}
		{\includegraphics[width=0.5\textwidth]{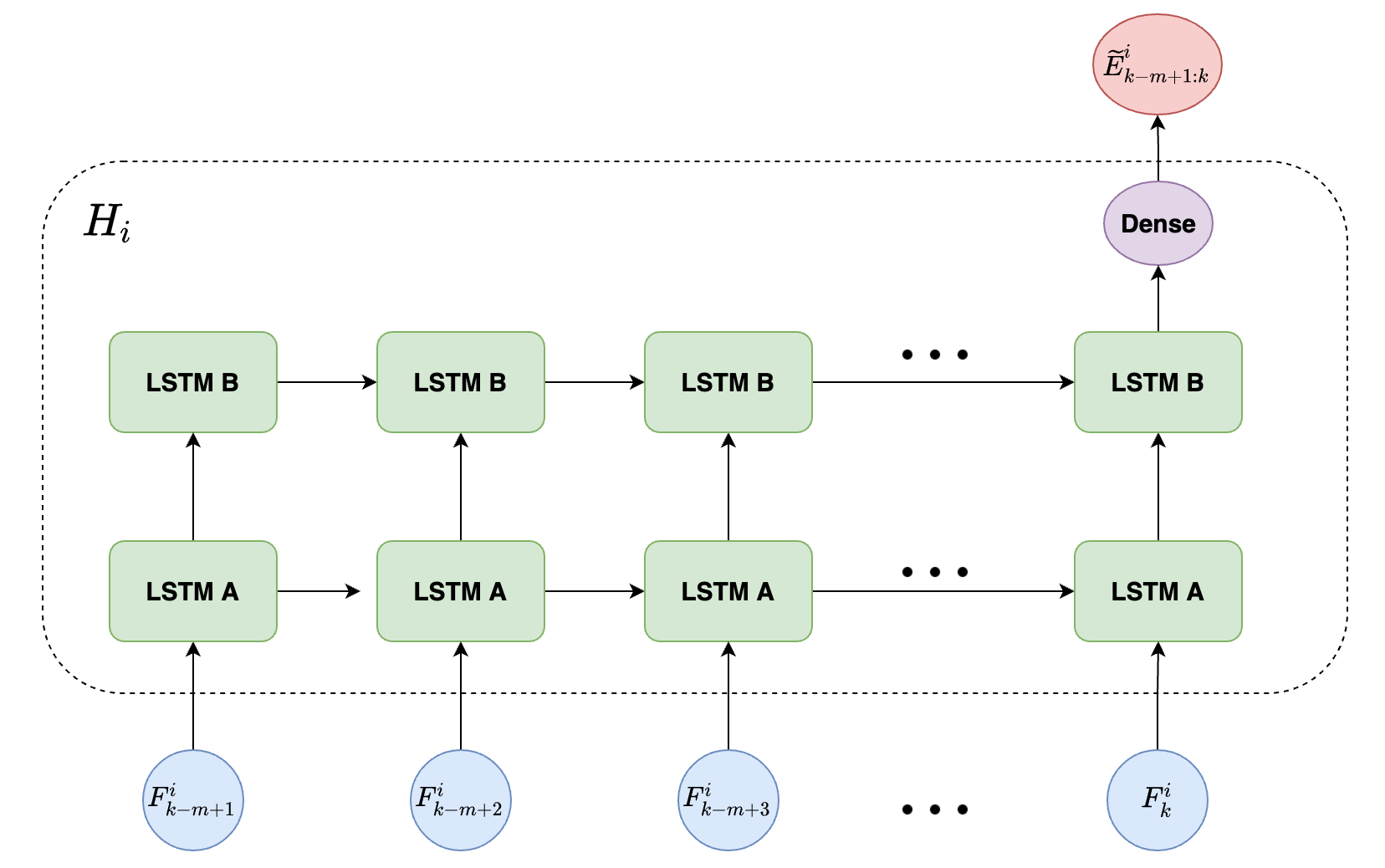}}
		\caption{The stacked LSTM architecture for each local model training.}
		\label{lstm}
	\end{center}
\end{figure}

Given the data from SUMO and Simulink for each vehicle, a local machine learning model can be trained as a prerequisite for federated learning. For this purpose, we adopted a stacked-LSTM (long short-term memory) architecture illustrated in Fig. \ref{lstm}, which includes two LSTM layers stacked together with dropout for regularization. 

At every given time, the input to the network is a sequential data which contains $m$ input points with each input being a feature vector with dimension $p$. The output of first LSTM layer returns a sequential data which is then taken as the input for the following LSTM layer. The output of the second LSTM layer is a single vector which captures information of all hidden cells in the stacked-LSTM network. This output passes through a fully connected dense layer with a hyperbolic tangent function as activation function to predict the overall energy consumption of the vehicle over the past $m$ steps. Here the tangent function is chosen as the model output can also be negative, i.e. regenerative energy. Please note that the stacked-LSTM architecture can include more LSTM layers as wish, here we use two LSTM layers for the architecture of our local model for baseline.

\subsection{Federated Averaging Learning Algorithm}

In this work, we adopted the well-known Federated Averaging (FedAvg) algorithm in \cite{mcmahan2017communication} to train a model and evaluate the performance of the model's accuracy. Now we briefly review the FedAvg algorithm in \cite{mcmahan2017communication} for readers’ conveniences. 


 \begin{algorithm}[htbp]
	\caption{Federated Averaging Algorithm (FedAvg)}
	\begin{algorithmic}[1]
		\State \textbf{Fed-BEV Server executes:}
		\Indent
		\State initialize $\omega_0$	    
		\For{each round $t=1,2,\ldots$}
		\State $m \leftarrow \max(C \cdot N, 1)$
		\State $S_t \leftarrow$ (random set of $m$ BEVs)		
		\For{each BEV $i \in S_t$ \textbf{in parallel}} 
		\State $\omega_{t+1}^i \leftarrow$ BEVUpdate$(i, \omega_t)$
		\EndFor 
		\State $\omega_{t+1} \leftarrow \frac{\sum_{i=1}^{N} \sharp D^i \omega_{t+1}^i}{\sum_{i=1}^{N} \sharp D^i}$
		\EndFor \\
		\EndIndent
		\State \textbf{BEVUpdate}$(i, \omega)$\textbf{:}  \quad \textit{// Run on BEV $i$}
	    \Indent
	    \State $\mathscr{B} \leftarrow$ (split $D^i$ into batches of size $B$)
		\For{each local epoch $p$ from $1$ to $P$}
		\For{batch $b \in \mathscr{B}$}
		\State $\omega \leftarrow \omega - \eta \nabla \ell(\omega; b)$
		\EndFor
		\EndFor
		\State return $\omega$ and $\sharp D^i$ to server
		\EndIndent
	\end{algorithmic}
	\label{fedavg}
\end{algorithm}

In Algorithm \ref{fedavg}, it can be seen that the implementation of the FedAvg algorithm consists of two parts including the server execution and BEV updates working in a nested and iterative manner. The algorithm starts by initializing weights for the global model at the server side. For every following iterative round $t$, the server selects a random subset of BEVs from the programme, and then each selected BEV conducts its local update (BEVUpdate). The local update essentially trains a local model through gradient descent using mini-batches split from each vehicle's local training dataset, where the function $\ell(w;b)$ is the training loss function parametrized by each local mini-batch $b$ with size $B$ and $\eta$ is the learning rate. The number of epochs for local model training, $P$, depends on the training dataset $\mathscr{B}$, which is clearly based on the dataset $D^i$ available at BEV $i$. At the end of each local model training session, a trained local model returns to the server for further processing. We note that the key difference between FedAvg and FedSGD is that FedAvg returns a trained local model $\omega$ to server while FedSGD returns the gradient information, i.e. $\nabla \ell$. Upon receiving the updated model parameters from BEVs in $S_t$, the Fed-BEV server aggregates the updated parameters together with the parameters from the rest of BEVs in the whole set proportionally, where the $i$'th proportion is defined as the fraction of the number of local training samples $\sharp D^i$ with respect to the overall training samples $\sum_{i=1}^{N} \sharp D^i$. 

\textbf{Comment on privacy:} Finally, since FedAvg requires each locally trained $\omega$ and $\sharp D^i$ to be returned back to the server at every iteration, this may impose security concerns to users as sensitive information contained in $\omega$ may be leaked as pointed out in some recent studies \cite{nasr2019comprehensive, bagdasaryan2020backdoor}. One way to deal with such a challenge is to apply encryption algorithms in implementation such as secure multiparty computation or homomorphic encryption \cite{dong2020eastfly} on $\omega$ and $\sharp D^i$. In this way, the averaging operation required by the 9th line of the Algorithm \ref{fedavg} can still be conducted securely without knowing the full information on both $\omega$ and $\sharp D^i$. As a result, an encrypted $\omega_{t+1}$, typically denoted by $[\omega_{t+1}]$, can be sent back to each BEV for following model iterations after local decryption.

\section{Simulation and Results} \label{Simulation}

In this section we discuss our system implementation using the proposed Fed-BEV in Fig. \ref{sys_architecture}. We first introduce the simulation setup and then discuss results in different settings. 

\subsection{Simulation Setup}

We assumed that there were 10 BEVs joined in the Fed-BEV programme as initial trial. The mobility of all BEVs was simulated in SUMO using the road networks imported from OpenStreetMap in the Dublin City, Ireland. The trips of all BEVs were generated using the shortest route method provided by SUMO for every given origin-destination pair. A driver's travelling behaviour was modelled using the car-following model (Krauss) in SUMO by default, but was parameterized by different acceleration and deceleration values to create various speed profiles for different trips. We assumed that all drivers obeyed the traffic rules, i.e. without speeding. In addition, the battery size was set to 40kWh for all vehicles referring to the Nissan Leaf model \footnote{https://www.nissan.ie/vehicles/new-vehicles/leaf/leaf-range-and-charging.html}. Each vehicle was also characterised by its weight, ranging from 2000kg to 2500kg uniformly. Finally, for a given trip, the elevation data of every GPS point along the journey was obtained through SUMO and was queried through the public API using the OpenTopoData \footnote{https://www.opentopodata.org/}.

\subsection{Network Setup}\label{networksetup}

A sequential stacked-LSTM model was required for each local model and it was constructed through Keras \footnote{https://keras.io/}. In our network setup, the input layer had 30 inputs and each input contained a vector with two dimensions, including both speed and elevation information at the given time point. By default, both LSTM layers included 50 hidden cells, and dropout was applied with a rate of 0.2 after each LSTM layer. The dense layer was applied for model output with hyperbolic tangent as the activation function. The output is a single value which captures the overall energy consumption of the vehicle over the past 30 seconds. The loss function was chosen as the mean absolute error (MAE) and the model was optimized using Adam optimizer. Each local dataset was split into 5 batches with 100 epochs for each local model training. By default, the learning rate for each local model was chosen to $1e^{-3}$ with the weight decay value equalling to $1e^{-5}$. Finally, the server was set to iterate 25 rounds for model aggregation.

\subsection{Simulation Results}

\begin{figure}[htbp]
	\begin{center}
		{\includegraphics[width=0.5\textwidth, height=3in]{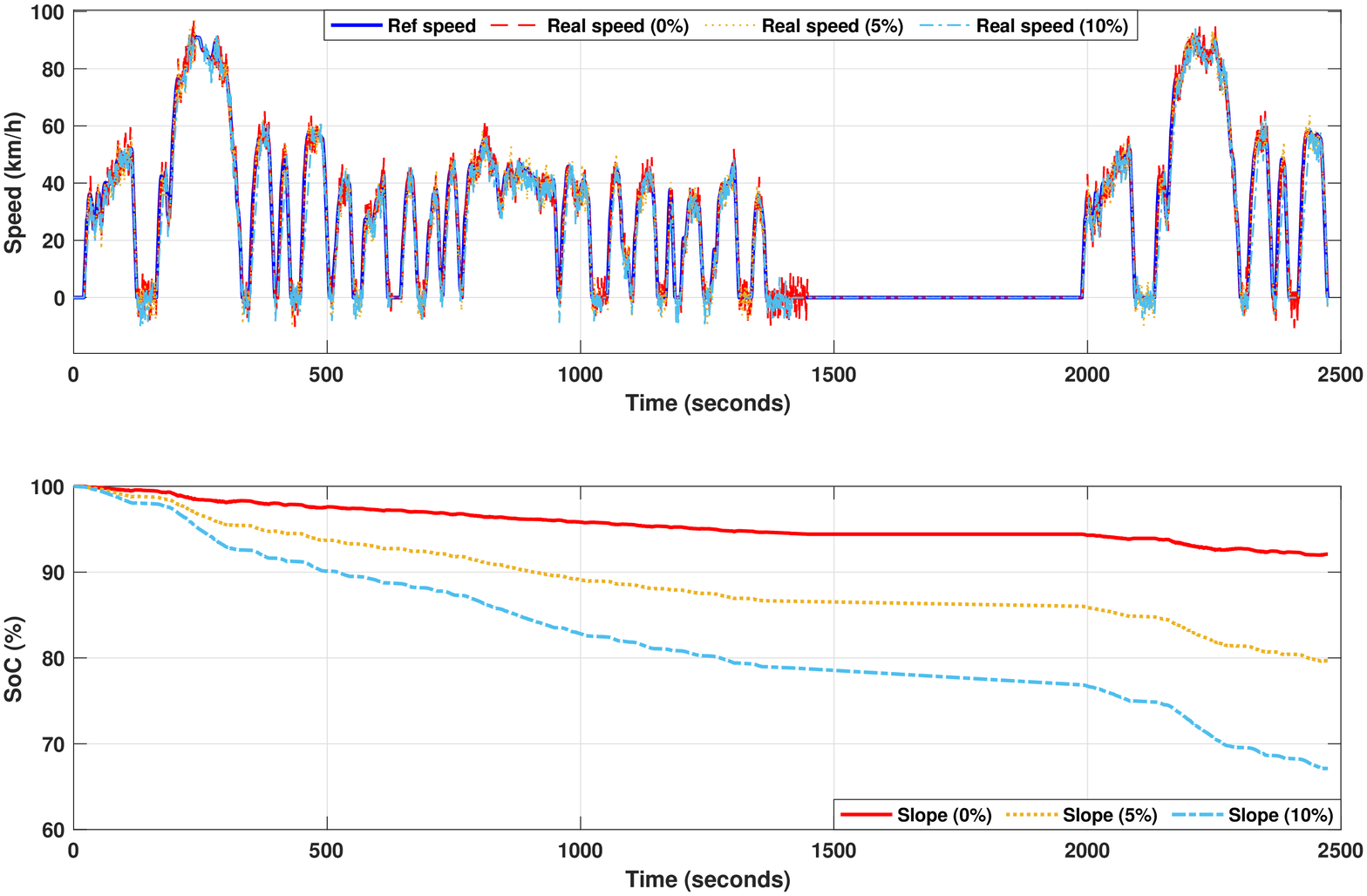}}
		\caption{Comparison of energy consumption of a BEV with respect to different road slopes.}
		\label{fed_result1}
	\end{center}
\end{figure}

\begin{figure}[htbp]
	\begin{center}
		{\includegraphics[width=0.5\textwidth]{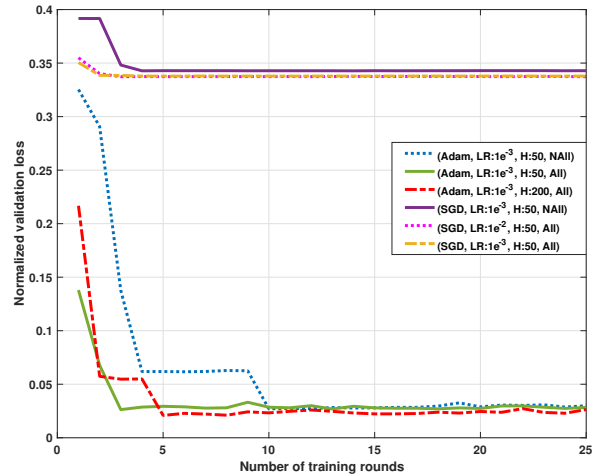}}
		\caption{Performance evaluation of FedAvg in different settings.}
		\label{fed_result2}
	\end{center}
\end{figure}

\begin{figure}[htbp]
	\begin{center}
		{\includegraphics[width=0.5\textwidth]{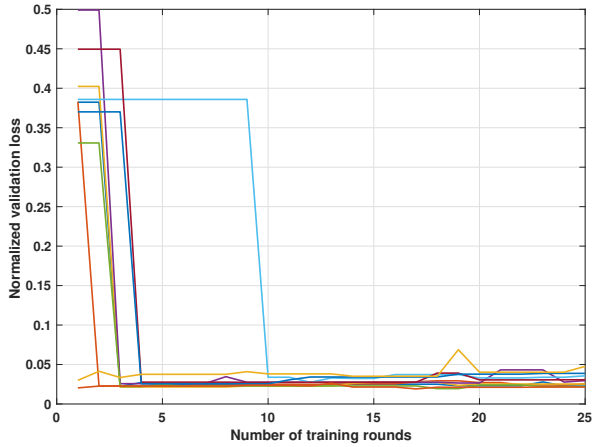}}
		\caption{Evaluation of FedAvg in local validation sets using the default setting.}
		\label{fed_result3}
	\end{center}
\end{figure}

\begin{figure}[htbp]
	\begin{center}
		{\includegraphics[width=0.5\textwidth]{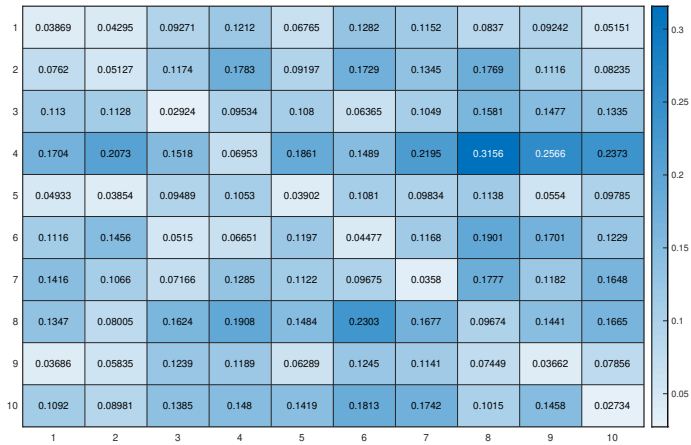}}
		\caption{Performance evaluation of each local model to other validation sets.}
		\label{fed_result4}
	\end{center}
\end{figure}

\begin{figure}[htbp]
	\begin{center}
		{\includegraphics[width=0.5\textwidth]{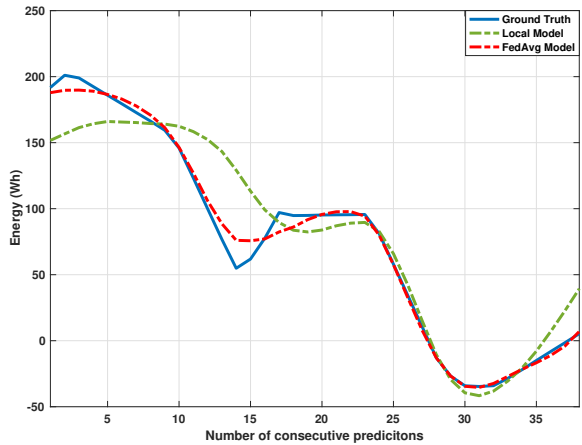}}
		\caption{Performance evaluation of the FedAvg algorithm and the local trained model with reference to ground truth of a testing energy consumption curve.}
		\label{fed_result5}
	\end{center}
\end{figure}

In this section, we conduct the experiments and present the simulation results under different setups in Figs \ref{fed_result1} - \ref{fed_result5}. More specifically, Fig. \ref{fed_result1} presents a sample result on how energy consumption patterns can be different for a given BEV along a specific city driving cycle (FTP-75) with respect to different road slopes. We note that the FTP-75 has been widely used for fuel economy testing of light-duty vehicles in the United States \cite{sluder2006estimate}. The driving cycle has been used in our simulations to calibrate the vehicular model for BEVs. The upper subplot of the Fig. \ref{fed_result1} shows the reference speed profile of the vehicle and the achieved real speed profiles of the vehicle in simulations under different road conditions. After calibration, all real speed profiles have been matched well with the reference profile. For illustrative purpose, the road slopes were assumed as constant values in this example with the lowest slope of 0\% and the highest slope of 10\%. The lower subplot further shows that the energy consumption pattern for a given BEV  (with weight 2000kg) on a higher slope road can be significantly different in comparison to the vehicle running on the flat road.

Our results in Fig. \ref{fed_result2} illustrate that how FedAvg algorithm can converge under different setups. First note that the default network parameter settings have been mentioned in Section \ref{networksetup}, here we intentionally adjusted some parameters to show the speed of algorithm convergence in other settings of interest. Clearly, the FedAvg converged to a much higher loss when the stochastic gradient descent (SGD) optimizer was applied instead of the Adam optimizer. Among the three better results with the Adam optimizer, the red curve converged best when 200 hidden cells were applied and all BEVs were involved in the model updates synchronically. The second best result comes to the green curve, and it can be seen that with only 50 hidden cells applied to both LSTM layers, the algorithm could converge faster with less computation efforts, and the final model accuracy was very promising when compared to the best result. Finally, the blue curve shows how FedAvg converged with selected number of BEVs in each round of iteration. This is the default setting of FedAvg as the algorithm does not need to be implemented synchronically while also achieving very competitive results in comparison to all others. 

In Fig. \ref{fed_result3}, it is shown that the normalized validation loss has been significantly reduced across all local validation sets after a few number of iterations. In order to further demonstrate the efficacy of the FedAvg in our application, we trained 10 local models for all BEVs using their respective local datasets based on the default stacked-LSTM architecture without federated learning. After that, each trained local model was applied to the validation sets belonging to all BEVs in order to calculate the validation loss matrix, which is depicted in Fig. \ref{fed_result4}. The results show that the generalization capabilities of all models were not quite comparable to the federated model, and this even applied to results of some models' local validation losses reflected over the diagonal line of the matrix. Our last results in Fig. \ref{fed_result5} compare the accuracy of the federated learning model and the local trained model for energy consumption prediction of a BEV with reference to the ground truth. It clearly shows that the FedAvg model makes more accurate prediction to the ground truth (blue curve) compared to the local trained model present in green curve.

\section{Conclusions} \label{Conclusion}

In this paper, an end-to-end federated learning framework has been proposed for modelling energy consumption of BEVs, namely Fed-BEV. The framework leverages the capabilities of the FedAvg algorithm to train a global model based on local models using stacked-LSTM architecture. The experimental results have shown that the FedAvg algorithm can easily enhance the prediction capabilities of local models through iterations in an asynchronized manner. We believe that the presented work is the first of its kind for energy modelling of BEVs using federated learning techniques. As part of our future work, we shall further integrate more features to the vehicular model, validate the effectiveness of the model using realistic datasets, explore and evaluate the efficacy of other centralized and decentralized federated learning paradigms for energy consumption modelling of BEVs.


\bibliographystyle{IEEEtran}
\bibliography{refs}             

\end{document}